%% file: main.tex
\documentclass{article}
\usepackage{styles/metrichand_arxiv,times}

\usepackage{hyperref,url,booktabs,amsmath,amsfonts,amssymb,xcolor,graphicx,array,multirow}
\usepackage{colortbl}
\usepackage{subcaption}
\usepackage{float}
\definecolor{metricFirst}{HTML}{FFD9D9}
\definecolor{metricSecond}{HTML}{FFEBCC}
\definecolor{metricThird}{HTML}{FFF8B8}
\definecolor{datasetHot}{HTML}{EAF1F7}
\definecolor{datasetArctic}{HTML}{F5F0E8}
\definecolor{referenceBlue}{rgb}{0.21,0.49,0.74}
\hypersetup{pdftitle={MetricHand: Geometry-Aware Flow Matching for 4D Hand Reconstruction in Metric World Space},pdfauthor={Yibo Bai, Likui Zhang, Jinglei Zhang, Zhengbin Pang, Xingshen Song, Xiaoqing Yin, Jinsheng Deng},colorlinks=true,citecolor=referenceBlue,linkcolor=referenceBlue,urlcolor=referenceBlue}
\newcommand{\method}{\textsc{MetricHand}}

\title{MetricHand: Geometry-Aware Flow Matching for 4D Hand Reconstruction in Metric World Space}
\author{%
Yibo Bai\textsuperscript{1}\quad
Likui Zhang\textsuperscript{2}\quad
Jinglei Zhang\textsuperscript{3}\quad
Zhengbin Pang\textsuperscript{1}\\[3pt]
Xingshen Song\textsuperscript{1,*}\quad
Xiaoqing Yin\textsuperscript{1}\quad
Jinsheng Deng\textsuperscript{4,*}\\[6pt]
{\small\textsuperscript{1}National University of Defense Technology}\\
{\small\textsuperscript{2}Sun Yat-sen University\quad
\textsuperscript{3}Shanghai Jiao Tong University}\\
{\small\textsuperscript{4}Academy of Military Sciences}\\[3pt]
{\small\textsuperscript{*}Corresponding authors.}}
\date{}
\renewenvironment{abstract}{%
  \par\addvspace{6pt}\begingroup\small
  \noindent\makebox[\linewidth]{\large\scshape Abstract}\par\nobreak
  \vspace{3pt}\noindent
}{\par\endgroup\addvspace{6pt}}
\begin{document}
\maketitle

\begingroup
\setlength{\intextsep}{8pt plus 2pt minus 2pt}
\begin{figure}[H]
\centering
\captionsetup{hypcap=false,skip=4pt,font=small}
\captionsetup[subfigure]{font=footnotesize,skip=1pt,justification=centering}
\begin{subfigure}[t]{0.49\linewidth}
\vspace{0pt}\includegraphics[width=\linewidth]{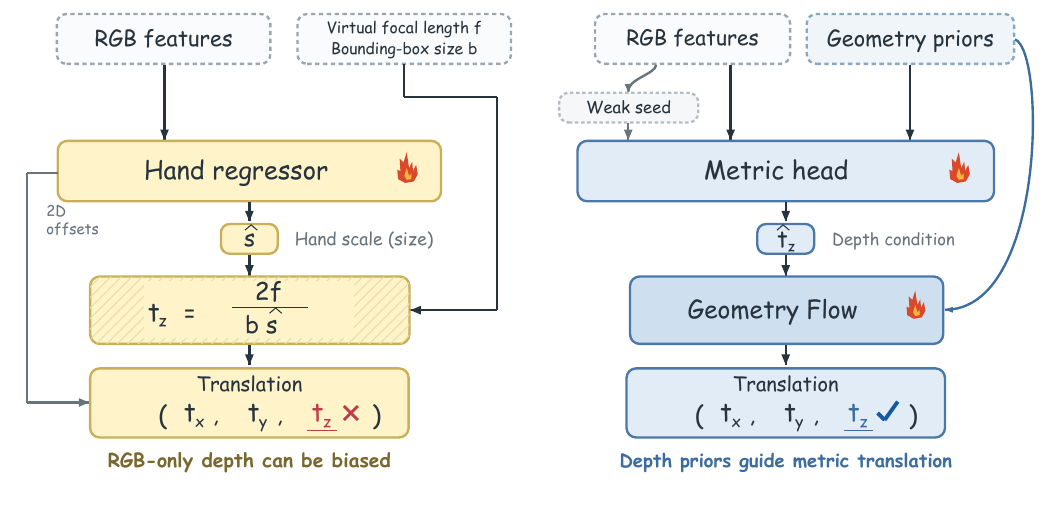}
\caption{Metric Translation Prediction}\label{fig:teaser-a}
\end{subfigure}\hfill
\begin{subfigure}[t]{0.49\linewidth}
\vspace{0pt}\includegraphics[width=\linewidth]{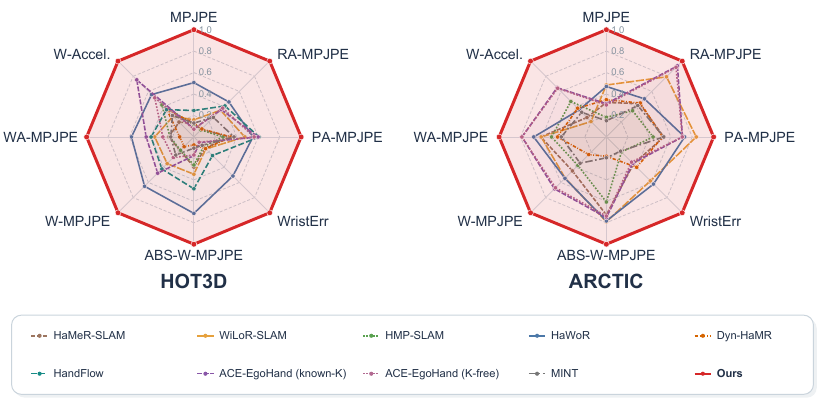}
\caption{Quantitative Comparisons on HOT3D and ARCTIC}\label{fig:teaser-b}
\end{subfigure}\par\vspace{2pt}
\begin{subfigure}[t]{\linewidth}
\includegraphics[width=\linewidth,trim=0 8bp 0 9bp,clip]{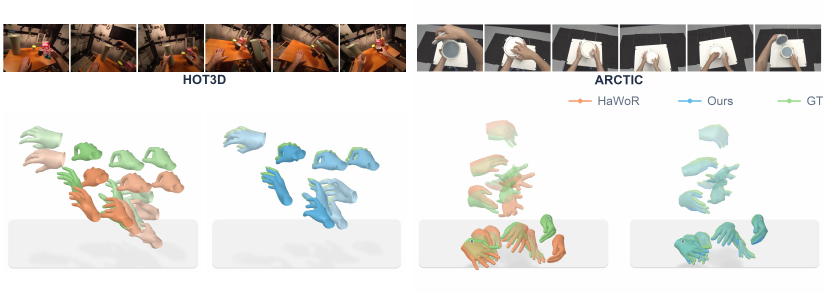}
\caption{HOT3D and ARCTIC hand motion}\label{fig:teaser-c}
\end{subfigure}
\caption{\textbf{MetricHand: geometry-guided 4D hand reconstruction.} 
(a) Weak-perspective projection can bias translation (left), whereas our method uses depth priors to guide metric translation estimation (right).
(b) Quantitative comparisons on HOT3D and ARCTIC under a shared evaluation protocol; higher is better.
(c) Qualitative world-space comparisons, with GT shown in green, ours in blue, and HaWoR in orange. Closer alignment with the green GT meshes indicates more accurate hand-pose reconstruction in the world coordinate system}
\label{fig:insight}
\label{fig:teaser}
\end{figure}
\endgroup

\begin{abstract}
Egocentric hand motion provides physical motion supervision for dexterous manipulation and vision-language-action pretraining. Reliable use of this supervision requires accurate and temporally coherent 4D hand reconstruction in metric world space. Existing methods typically lift framewise estimates produced under weak perspective into world coordinates. By ignoring joint-dependent depth variation at close range, this approximation couples errors in translation, orientation, and articulation that persist after world lifting and worsen under occlusion.
We present \method{}, a geometry-aware flow matching framework for 4D hand reconstruction in metric world space. \method{} encodes depth priors from hand--object interaction regions into joint-associated representations that preserve local spatial structure and metric scale. Fused with visual features, this geometry guides coarse metric estimation and conditions every step of temporal generation. The model jointly recovers world-space translation, global orientation, MANO articulation, and shape, while structural priors and temporal context support estimation under partial occlusion.
\method{} reduces camera-space MPJPE by 49.3\% and 52.9\%, absolute world-space MPJPE by 28.7\% and 21.7\%, and wrist localization error by 48.5\% and 37.9\% relative to HaWoR, respectively. World-space acceleration error decreases by 44.2\% and 64.3\%, demonstrating improved temporal fidelity alongside more accurate absolute localization.
Our project page is available at \url{https://bieber1205.github.io/MetricHand/}.
\end{abstract}

\section{Introduction}
\label{sec:introduction}
Hand motion lies at the center of how humans interact with the physical world. In egocentric video, wrist position and orientation specify end-effector targets, while local hand pose defines the articulated configuration that can be transferred to a robot hand through retargeting. These states turn human manipulation videos into scalable physical motion supervision for dexterous manipulation learning and vision-language-action (VLA) pretraining~\citep{kareer2025egomimic,hoque2026egodex,yang2025egovla}. Localization errors can shift hand--object interaction targets, and temporal jitter can propagate into generated trajectories. These applications therefore require accurate and temporally coherent 4D hand motion in a common world coordinate system with metric scale.

Existing egocentric 4D hand reconstruction commonly combines two stages. Image-based methods such as HaMeR and WiLoR recover detailed hand observations and visual features from individual frames~\citep{pavlakos2023hamer,potamias2025wilor}. Temporal methods such as HaWoR and HandFlow combine these estimates with camera motion to reconstruct world-space trajectories~\citep{zhang2025hawor,xu2026handflow}. These approaches improve temporal consistency under moving cameras, but they still depend on the image frontend to establish the geometric state of the hand in each frame. When that estimate is biased or the hand is occluded, temporal processing must recover motion from incomplete and potentially distorted evidence.

Metric hand reconstruction requires resolving the depth ambiguity left by weak-perspective estimates and camera-normalized rays. HandFlow~\citep{xu2026handflow} conditions flow matching on visual features and skeletal rays, with detection confidence controlling observation strength. StableHand~\citep{zeng2026stablehand} uses predicted observation quality to adapt the flow path and initialization, alongside hand, camera, and scene-geometry conditions. Our focus is the spatial organization of geometric evidence: associating metric depth and nearby hand--object surfaces with individual joints so that the generator can use these cues for both placement and articulation.

Our key insight is that a topology-aware representation of hand--object interaction (HOI) geometry can serve as a shared condition throughout 4D hand reconstruction. Local surface structure provides cues for recovering hand articulation under occlusion, while metric depth guides hand localization in world coordinates. Integrating this geometric evidence with temporal memory further supports motion consistency across frames and trajectory continuity across windows. Based on this insight, \method{} fuses RGB and estimated depth for coarse metric hand estimation, then samples appearance and surface geometry around 21 projected joint anchors. Quality weighting and temporal memory organize these features across windows, and the resulting conditions guide every flow update to jointly recover world-space translation, global orientation, articulation, and shape. Our geometry-conditioning ablations show that maintaining this evidence throughout generation improves reconstruction over using depth only for coarse estimation or supplying geometry only at the first flow step (Table~\ref{tab:component_multidataset}).

Under a shared protocol on HOT3D and ARCTIC, the evaluated configuration achieves the lowest wrist localization error and absolute camera- and world-space joint errors among the reevaluated methods. Relative to HaWoR, absolute world-space MPJPE decreases by 28.7\% and 21.7\%, and world-space acceleration error by 44.2\% and 64.3\%, respectively. These results demonstrate improvements in absolute localization and motion fidelity (Fig.~\ref{fig:teaser}). Component and condition ablations assess geometric representation, flow generation, and temporal memory.

Our contributions are:
\begin{itemize}
    \item We propose \method{}, which associates the spatial structure and metric scale of HOI depth priors with hand topology, providing shared geometric conditions for metric localization and occlusion-robust pose estimation.
    \item We formulate geometry-aware temporal flow matching that combines quality-aware conditions with overlapping memory to jointly generate translation, global orientation, articulation, and shape.
    \item We achieve state-of-the-art hand reconstruction on HOT3D, ARCTIC, DexYCB, and HO3D, as well as on unseen real-world HOI4D videos under zero-shot evaluation.
\end{itemize}

\section{Related Work}
\label{sec:related_work}
\paragraph{Image-based Hand Reconstruction.}
MANO represents hand shape and articulation with a differentiable parametric mesh~\citep{romero2017embodied}. Transformer-based models such as HaMeR and WiLoR recover detailed hands from images~\citep{pavlakos2023hamer,potamias2025wilor}; Hamba and HandOS explore alternative spatial reasoning and localization architectures~\citep{dong2024hamba,chen2025handos}. These models provide visual features and articulation priors. Recovering metric motion additionally requires geometric evidence that connects the reconstructed pose to its physical placement.

\paragraph{Hand Motion Estimation and Generation.}
Temporal methods combine image evidence with motion priors to reconstruct hand trajectories. HMP learns a hand-motion prior~\citep{duran2024hmp}, while Dyn-HaMR and HaWoR recover motion under moving cameras~\citep{Dynhamr,zhang2025hawor}. ViDiHand adapts video-diffusion representations for hand reconstruction~\citep{wang2026vidihand}. ACE-Ego-Hand uses a deterministic clean-latent encoder and a bidirectional spatiotemporal decoder to recover bimanual trajectories under occlusion and out-of-sight gaps~\citep{liu2026aceegohand}. MINT jointly predicts camera motion, field of view, camera-frame hand states, and observability from a shared video representation, using large-scale pipeline supervision~\citep{zhu2026mint}. HandFlow formulates hand recovery with flow matching and confidence-aware observation conditioning~\citep{xu2026handflow,lipman2023flow}. Our focus is the explicit geometric evidence supplied to temporal generation: topology-aware sampling associates HOI appearance, metric depth, and local surface structure with hand joints, and the resulting conditions remain available throughout the flow process.

\paragraph{Metric Depth Estimation.}
Monocular geometry models such as Metric3D, Metric3D v2, and MoGe-2 recover depth and metric-scale cues from a single image~\citep{metric3d,metric3dv2,moge2}. Their spatial predictions complement the appearance features of a normalized hand crop. BLADE uses depth to address close-range perspective distortion in body reconstruction~\citep{wang2025blade}. For articulated hand motion, depth provides both absolute placement cues and the local layout of the interaction region. We fuse this evidence with RGB features, sample it around hand joints, and preserve it across time as a shared condition for localization and temporally coherent pose estimation. Camera-motion estimation supplies the reference needed to express this geometry in world coordinates~\citep{teed2021droid,zhang2025hawor}.

\section{Method}
\label{sec:method}
\begin{figure}[t]
\centering
\includegraphics[width=\linewidth]{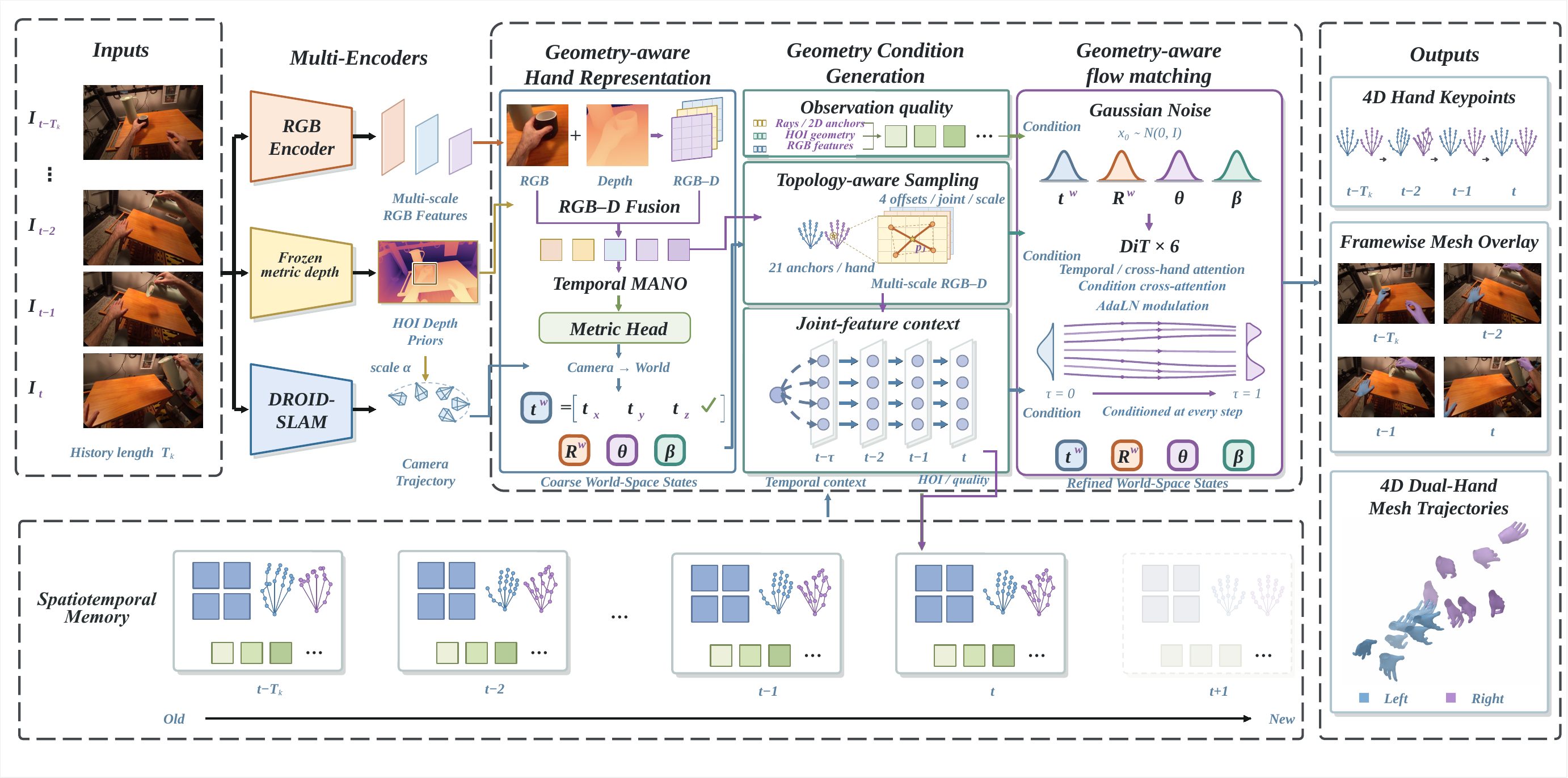}
\caption{\textbf{Overview of MetricHand.}
Depth priors supply metric placement cues and local hand--object surface geometry. RGB--D fusion supports coarse hand estimation, while topology-aware sampling associates multi-scale evidence with projected joint anchors. Observation quality and a spatiotemporal memory queue preserve reliable joint-associated context across windows. The resulting conditions are supplied to every flow-matching step to jointly recover world-space wrist translation, global orientation, articulation, and shape. Camera motion defines the metric world reference.}
\label{fig:architecture}
\end{figure}

\paragraph{Overview.}
\method{} uses interaction geometry to connect metric hand placement with temporal pose recovery (Fig.~\ref{fig:architecture}). RGB features and depth priors first produce coarse hand estimates, which are expressed in a metric world frame using the estimated camera trajectory (Sec.~\ref{sec:rgbd}). Topology-aware sampling then gathers appearance and surface geometry around the projected hand joints. Observation quality and cross-window memory organize this evidence into joint-associated conditions (Sec.~\ref{sec:condition_generation}). A conditional flow model reads these conditions at every integration step to jointly recover world-space translation, global orientation, articulation, and shape (Sec.~\ref{sec:flow}).

\subsection{Preliminaries}
\label{sec:geometry}
\paragraph{Problem Definition.}
Given a monocular video $\{I_t\}_{t=1}^{T}$ and intrinsics $K_t$, we use estimated depth $D_t$ and camera-to-world poses $\mathcal T_t^{WC}=(R_t^{WC},c_t^{WC})$ to reconstruct both hands $h\in\{L,R\}$ in a common metric world frame.

\paragraph{4D Hand Motion Representation.}
We represent each hand with MANO~\citep{romero2017embodied}. At frame $t$, $\mathbf t_{t,h}^{W}$ is the global translation defined by the physical wrist, $R_{t,h}^{W}$ the global orientation, $\theta_{t,h}$ the 15 local joint rotations, and $\beta_h\in\mathbb R^{10}$ the hand shape. Translation, orientation, and articulation vary by frame, while shape is shared within a window. Separate observation and annotation masks control conditioning and supervision.

\subsection{Geometry-Aware Hand Representation}
\label{sec:rgbd}

\paragraph{Multi-Encoders.}
Following image-based hand reconstruction~\citep{pavlakos2023hamer,potamias2025wilor}, a trainable WiLoR-initialized encoder extracts hand--object appearance features $F^I$. A frozen monocular geometry model supplies full-frame metric depth~\citep{metric3dv2,moge2}, and frozen DROID-SLAM estimates relative camera motion~\citep{teed2021droid}. Depth is aligned to each hand crop with the RGB transform. RGB--D fusion, Temporal MANO, and the Metric Head are trained jointly, while the depth and SLAM networks remain fixed.

\paragraph{RGB--D Fusion.}
The representation combines dense depth injection with joint-associated geometric aggregation. The depth encoder processes normalized log depth, validity, and spatial gradients to preserve local surface layout, while absolute-depth statistics retain metric context. Depth features are aligned with the RGB feature map and injected through a spatially gated residual:
\begin{equation}
 F^{RD}=F^I+\operatorname{sigmoid}
 \big(a+g(\operatorname{GN}(F^D))\big)
 \odot\operatorname{GN}(F^D),
 \label{eq:fusion_main}
\end{equation}
where $g$ is a learned projection, $a$ a learned bias, and GN denotes group normalization. Temporal MANO uses $F^{RD}$ to predict coarse orientation, articulation, camera translation, and shared shape. The fused map also supplies the multi-scale evidence for topology-aware sampling in Sec.~\ref{sec:topology}. Projected coarse joints index this evidence by anatomical identity, allowing the subsequent flow model to use local hand--object geometry together with metric placement cues.

\paragraph{Metric Head.}
Crop scale and weak perspective leave translation ambiguous at close range~\citep{wang2025blade}. Our Metric Head uses fused features, crop geometry, and absolute depth statistics to refine the coarse MANO translation $\widetilde{\mathbf t}^{C}$ in viewing-ray and log-depth coordinates:
\begin{equation}
 \widehat r=\widetilde r+0.25\delta_{xy},\qquad
 \widehat\zeta=\operatorname{clip}
 (\widetilde\zeta+\delta_z,\log z_{\min},\log z_{\max}),\qquad
 \widehat{\mathbf t}_{\mathrm{MANO}}^C=e^{\widehat\zeta}
 [\widehat r_x,\widehat r_y,1]^\top,
 \label{eq:metric_seed_main}
\end{equation}
where $\delta=\tanh H_{\mathrm{metric}}(\cdot)$, $\widetilde r=\widetilde{\mathbf t}^{C}_{xy}/\max(\widetilde t_z^C,\epsilon)$, and $\widetilde\zeta$ is clipped log depth. After adding the MANO root offset, the physical wrist and orientation are transformed by the camera-to-world pose. We mask dynamic hand--object regions and recover DROID-SLAM scale by matching its depth to the metric prior on valid static pixels, following metric-scaled world lifting~\citep{zhang2025hawor}. The coarse states and later conditions therefore share one metric world frame.

\subsection{Geometry Condition Generation}
\label{sec:condition_generation}
The condition generator turns crop features into structured evidence. Joint anchors align RGB--D features with hand topology, quality prediction suppresses unreliable content, and a spatiotemporal queue links joint identities across windows. The output preserves local hand--object geometry, global context, and temporal correspondence for the flow model.

\paragraph{Topology-Aware Sampling.}
\label{sec:topology}
We construct an HOI representation indexed by anatomical joints. The 21 projected coarse MANO joints provide stable anchors for gathering appearance and geometry from nearby hand and object surfaces. Each joint query predicts offsets on multi-scale RGB--D features, following deformable sampling~\citep{zhu2021deformabledetr}. Content-based attention is augmented with image-distance, log-depth, and backprojected surface-distance biases, so each joint token retains the spatial organization of its supporting evidence. Object surfaces provide context for pose recovery under occlusion rather than joint-position targets. The query predicts $u_{j,k}=u_j+\Delta u_{j,k}$. Backprojecting valid depth as $P^C(u)=D(u)K^{-1}\bar u$, we define a geometric bias:
\begin{equation}
 \begin{aligned}
 b_{j,k}={}&-a_u\|\chi(u_{j,k})-\chi(u_j)\|_2
 -a_z|\log D(u_{j,k})-\log D(u_j)|\\
 &-a_P\frac{\|P^C(u_{j,k})-P^C(u_j)\|_2}{\max(D(u_0),\epsilon)}.
 \end{aligned}
 \label{eq:geometry_bias}
\end{equation}
Here $\bar u$ is the homogeneous pixel coordinate, $\chi$ maps pixels to normalized image coordinates, $u_0$ is the wrist anchor, and $a_u,a_z,a_P$ are nonnegative learned coefficients. Depth-dependent terms use valid samples only. Combining this bias with the content logit $a_{j,k}$ gives
\begin{equation}
 w_{j,k}=\operatorname{softmax}_{k}(a_{j,k}+b_{j,k}),\qquad
 f_j=z_j+\nu_j W_O\sum_k w_{j,k}W_VF^{RD}(u_{j,k}),
 \label{eq:anchor_sampling}
\end{equation}
where $\nu_j$ gates the update by valid geometric support. Each $f_j$ associates appearance and surface geometry with one joint. Mask pooling, box encoding, and HOI pooling provide complementary hand-level and interaction context.

\paragraph{Observation Quality.}
\label{sec:conditioning}
Inspired by confidence-aware conditioning in HandFlow~\citep{xu2026handflow}, we separate availability from reliability. Hand- and joint-level heads combine RGB--D descriptors with detection confidence, depth support, camera features, and the coarse state into $z_i$. For condition slot $i$, a learned quality score weights its content $c_i$ against a placeholder:
\begin{equation}
 q_i=Q_\phi(z_i)\in[0,1],
 \quad \widetilde c_i=o_iq_i c_i+(1-o_iq_i)e_{\mathrm{placeholder}}+e_i^{\mathrm{id}}.
 \label{eq:quality_condition}
\end{equation}
Here $o_i$ indicates availability, while $e_i^{\mathrm{id}}$ preserves source, time, hand, and joint identity. Hand-level scores weight regional tokens and joint-level scores weight joint tokens; modality masks retain source validity. Quality scores are learned jointly through the reconstruction objectives, without explicit quality targets or an auxiliary loss.

\paragraph{Spatiotemporal Memory Queue.}
\label{sec:memory}
The queue carries topology-aware HOI features across windows. Each joint token $g_{t,h,j}$ is obtained from the sampled descriptor $f_j$ in Eq.~\ref{eq:anchor_sampling}, with observation quality and identity encoding applied as in Eq.~\ref{eq:quality_condition}. It therefore retains local appearance and surface geometry alongside its timestamp, hand identity, and joint identity. Spatial attention exchanges evidence between joints within a frame, and temporal attention follows the same anatomical joint through the current window and queued features. This correspondence allows a partially occluded joint to retrieve contextual evidence from visible joints and preceding observations. We maintain the FIFO queue $\mathcal Q_n$ indexed by timestamp, hand, and joint identity~\citep{zhang2025hawor,xu2026handflow}:
\begin{equation}
 \begin{aligned}
 g^S_{t,h,j}&=g_{t,h,j}+\operatorname{Attn}
 \big(g_{t,h,j},\{g_{t,h',k}\}_{h',k}\big),\\
 g^{ST}_{t,h,j}&=g^S_{t,h,j}+\operatorname{Attn}
 \big(g^S_{t,h,j},\{g^S_{s,h,j}\}_{s\in\mathcal W_n\cup\mathcal T(\mathcal Q_n)}\big).
 \end{aligned}
 \label{eq:queue_attention}
\end{equation}
Here $\mathcal W_n$ is the current window; $\mathcal T(\mathcal Q_n)$ contains only earlier timestamps, so the two sets are disjoint. Invalid observations are masked as keys, while placeholder queries can retrieve valid context. As the window advances, departed timestamps enter the history with detached features and predicted states; duplicate entries are merged and the oldest are evicted. Predicted states retain their observation-validity labels.

\paragraph{Condition Encoding.}
The encoder combines camera-ray cues, 2D skeletal anchors, topology-aware HOI features, and observation quality into $C_n$, projecting current and queued evidence to a shared width while retaining source, time, hand, and joint identities. $C_n$ supplies cross-attention, while its pooled representation $\bar C_n$ modulates the flow blocks.

\subsection{Geometry-Aware Flow Matching}
\label{sec:flow}
\label{sec:training}
Direct framewise regression lacks a unified temporal update when local evidence is biased or partially occluded. Building on flow matching and its use for generative hand recovery~\citep{lipman2023flow,xu2026handflow}, we map a two-hand window to motion tokens with component, time, and hand identities. Rotations use continuous 6D coordinates~\citep{zhou2019continuity}, translation is normalized, and shape is shared within the window. The velocity field $v_\phi(x_\tau,\tau,C_n)$ transports Gaussian noise toward the joint state distribution, with the coarse estimate supplied through $C_n$ rather than used as the initial sample.

\paragraph{Condition-Guided Generation.}
DiT-style blocks~\citep{peebles2023dit,xu2026handflow} couple motion tokens across state components, time, and hands through self-attention, and retrieve $C_n$ through cross-attention. Flow time $\tau$ and pooled context $\bar C_n$ modulate normalization and residual gates. Every velocity evaluation reads $C_n$: depth guides placement, while joint-associated HOI and memory features guide articulation and temporal consistency.

\paragraph{Flow Integration.}
Following the rectified-flow formulation used by HandFlow~\citep{lipman2023flow,xu2026handflow}, we connect Gaussian motion $x_0$ and normalized ground truth $x_1$ with a linear path:
\begin{equation}
 \begin{aligned}
 x_\tau&=(1-\tau)x_0+\tau x_1,\qquad u_\tau=x_1-x_0,\\
 \mathcal L_{\mathrm{FM}}
 &=\sum_k
 \mathbb E\!\left[
 \|v_\phi^k(x_\tau,\tau,C_n)-u_\tau^k\|^2
 \right]_{\mathrm{valid},k}.
 \end{aligned}
 \label{eq:flow}
\end{equation}
Here $k$ indexes shape, global orientation, articulation, and translation. Each component is averaged over its valid annotations before summation, preventing high-dimensional articulation from dominating the other states. Following HandFlow's few-step setting, inference integrates $dx_\tau/d\tau=v_\phi(x_\tau,\tau,C_n)$ with three Euler steps~\citep{xu2026handflow}. For long videos, repeated global timestamps share their initial noise as in HandFlow~\citep{xu2026handflow}; our implementation then blends window endpoints with triangular weights.

\begin{samepage}
\paragraph{Training Objectives.}
We supervise the MANO-decoded endpoint $\widehat x_1=x_\tau+(1-\tau)v_\phi(x_\tau,\tau,C_n)$. For the physical wrist residual $\mathbf e^W_{t,h}=\widehat{\mathbf t}^W_{t,h}-\mathbf t^W_{t,h}$ in metres, temporal and world-space translation supervision are
\begin{equation}
 \begin{aligned}
 \mathcal L_{\mathrm{temp}}
 &=\frac{1}{N_{\mathrm{temp}}}\sum_{t=2}^{L-1}\sum_{h\in\{L,R\}}
 m^{\mathrm{temp}}_{t,h}\,\rho_{\delta_{\mathrm{temp}}}\!\left(
 \mathbf e^W_{t+1,h}-2\mathbf e^W_{t,h}+\mathbf e^W_{t-1,h}\right),\\
 \mathcal L_{\mathrm{trans}}
 &=\frac{1}{N_{\mathrm{trans}}}\sum_{t=1}^{L}\sum_{h\in\{L,R\}}
 m_{t,h}\,\rho_{\delta_{\mathrm{trans}}}\!\left(
 \widehat{\mathbf t}^W_{t,h}-\mathbf t^W_{t,h}\right),
 \end{aligned}
 \label{eq:wrist_losses}
\end{equation}
Here $\rho_\delta$ is coordinate-averaged Smooth L1; $m$ masks valid wrists or consecutive triplets, and $N$ counts them. The temporal term matches target accelerations. The total objective is
\begin{equation}
 \mathcal L=\mathcal L_{\mathrm{FM}}
 +\mathcal L_{\mathrm{metric}}+\mathcal L_{\mathrm{pose}}
 +\lambda_{\mathrm{temp}}\mathcal L_{\mathrm{temp}},
 \label{eq:objective}
\end{equation}
The metric group includes $\lambda_{\mathrm{trans}}\mathcal L_{\mathrm{trans}}$.
\end{samepage}

\section{Experiments}
\label{sec:experiments}
\input{tables/crossdit3_main}
\subsection{Experimental Setup}
\label{sec:setup}
\paragraph{Datasets.}
We follow the observed-track evaluation convention of HaWoR~\citep{zhang2025hawor} and compare methods on the same valid observed frames of HOT3D and ARCTIC~\citep{banerjee2025hot3d,fan2023arctic}, both captured with moving egocentric cameras.

\paragraph{Metrics.}
We report camera MPJPE and absolute world MPJPE (ABS-W-MPJPE) for joint localization. WristErr is the Euclidean error of the single physical wrist point in camera coordinates. RA-MPJPE subtracts the wrist, whereas PA-MPJPE fits a per-frame similarity transform. World W-MPJPE and WA-MPJPE use blockwise similarity alignment fitted to the first two frames or the full block, respectively~\citep{zhang2025hawor}. Position errors are in millimetres. World acceleration error, in m/s$^2$, measures disagreement with ground-truth joint dynamics on the original timeline. The world reference is established by rigid camera alignment, with no further hand alignment for ABS-W-MPJPE.

\subsection{Comparison with State of the Art}
\label{sec:main_results}
\label{sec:temporal_results}

\paragraph{Baselines.}
We compare against SLAM-lifted HaMeR, WiLoR, and HMP~\citep{pavlakos2023hamer,potamias2025wilor,duran2024hmp}, and video-based methods including Dyn-HaMR, HaWoR, HandFlow, ACE-Ego-Hand, and MINT~\citep{Dynhamr,zhang2025hawor,xu2026handflow,liu2026aceegohand,zhu2026mint}. Dyn-HaMR retains its native optimized world trajectory.

\paragraph{Quantitative Results.}
Table~\ref{tab:main_results} shows that \method{} achieves the lowest absolute joint and wrist localization errors among the reevaluated methods on both datasets. We analyze the results in terms of metric localization, pose accuracy, and temporal consistency.

For \emph{metric localization}, ABS-W-MPJPE reaches 20.58 and 120.86~mm on HOT3D and ARCTIC, reducing HaWoR's errors by 28.7\% and 21.7\%, respectively. Wrist localization error also decreases by 48.5\% and 37.9\%, indicating more accurate physical placement of the reconstructed hands.

For \emph{pose accuracy}, PA-MPJPE is 3.18 and 5.18~mm on the two datasets, the lowest values in the tables. Camera-space MPJPE reaches 8.60 and 17.54~mm, showing improvements in both similarity-aligned pose and absolute hand geometry.

For \emph{temporal consistency}, world-space acceleration error is 3.11 and 4.39~m/s$^2$, improving over HaWoR by 44.2\% and 64.3\%, respectively. These results show closer agreement with ground-truth motion dynamics alongside improved localization.

\begin{figure}[t]
\centering
\includegraphics[width=\linewidth]{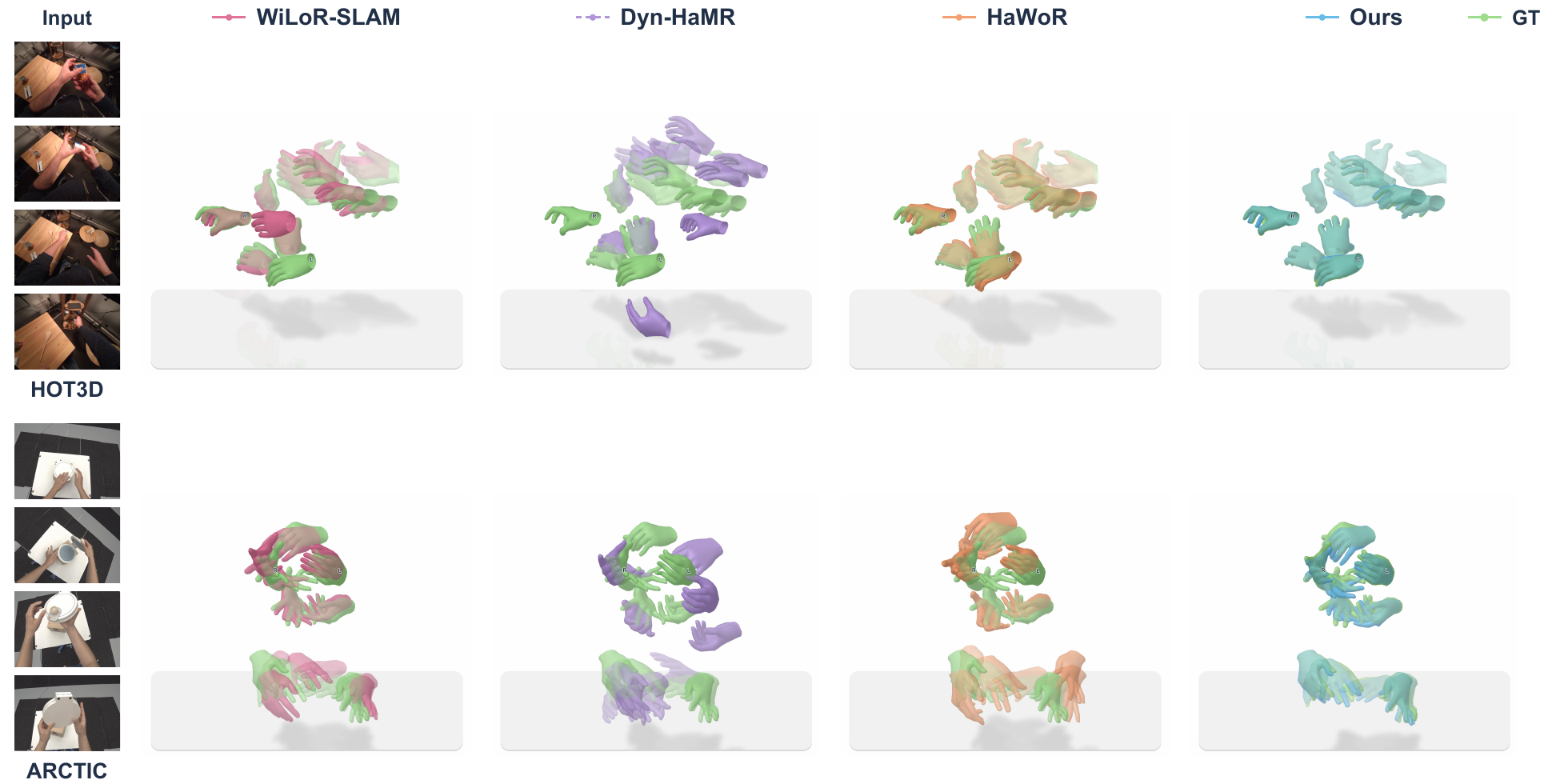}
\caption{\textbf{Qualitative comparison of world-space hand trajectories on HOT3D and ARCTIC.} Input frames run from top to bottom within each row. Green meshes show ground truth; pink, purple, orange, and blue show WiLoR-SLAM, Dyn-HaMR, HaWoR, and MetricHand, respectively. Closer overlap with the green GT meshes indicates higher hand-pose accuracy along the displayed world-space trajectories.}
\label{fig:qualitative}
\end{figure}

\paragraph{Qualitative Comparison.}
Figure~\ref{fig:qualitative} compares the reconstructed motion during object manipulation. Read the RGB frames from top to bottom, then follow the overlaid hand states from translucent early poses to opaque late poses. Each method is shown from the same view with green GT meshes overlaid. MetricHand follows the changing palm orientation and finger configuration in both examples. In HOT3D, its palm and finger contours remain close to GT as the hands move around the object; in ARCTIC, it preserves the transition between open and flexed fingers. These examples illustrate the articulated geometry recovered by the RGB-D-conditioned flow.

\subsection{Ablation Study}
\label{sec:ablations}
We emphasize camera-space MPJPE and evaluate every ablation with the same SLAM trajectory. Table~\ref{tab:ablation_summary} compares depth representations; Table~\ref{tab:component_multidataset} reports component and geometry-conditioning ablations on HOT3D and ARCTIC.

\input{tables/ablation_summary}
\input{tables/component_multidataset}

\paragraph{Depth Representation.}
Our RGB-D baselines concatenate image and depth features or fuse them through cross-attention before direct regression. On ARCTIC, these reduce RGB-only MPJPE from 36.34 to 31.65 and 29.64~mm. Topology-aware sampling associates HOI geometry with hand joints and reaches 19.98~mm, improving on the two fusion baselines by 36.9\% and 32.6\%. In the full-model ablation, removing topology-aware sampling raises MPJPE from 17.54 to 38.91~mm. Both comparisons support the value of structured geometric conditions.

\paragraph{Flow Matching.}
Replacing flow with direct regression retains all conditions, the queue, and the window length. Flow reduces ARCTIC MPJPE from 19.98 to 17.54~mm (12.2\%) and acceleration error from 7.93 to 4.39~m/s$^2$ (44.6\%). HOT3D shows the same trend. Flow therefore improves pose accuracy and agreement with ground-truth dynamics under the same geometric conditions.

\paragraph{Memory Queue.}
Removing history raises MPJPE from 8.60 to 15.42~mm on HOT3D and from 17.54 to 25.67~mm on ARCTIC. ARCTIC acceleration error also rises from 4.39 to 6.31~m/s$^2$, showing that cross-window context improves both reconstruction and temporal consistency.

\paragraph{HOI Context and Quality.}
On ARCTIC, removing regional HOI context or quality weighting raises MPJPE to 32.42 or 25.35~mm, while PA-MPJPE changes from 5.18 to 5.49 or 5.40~mm. Their larger effects on unaligned error indicate a stronger contribution to metric placement than to similarity-aligned pose. Every component removal worsens all reported metrics on both datasets.

\section{Conclusions and Limitations}
\label{sec:conclusion}
\textbf{Conclusion.} We presented \method{}, a geometry-aware flow matching framework for reconstructing 4D hand motion in metric world space. Our key insight is to use hand--object interaction geometry as a shared condition for metric localization and occlusion-robust pose estimation. HOI depth and visual features guide coarse metric estimation and form joint-associated conditions that retain spatial structure and metric scale. Quality-aware conditioning and temporal memory maintain this evidence throughout flow generation, jointly recovering world-space translation, global orientation, articulation, and shape. Experiments on HOT3D and ARCTIC demonstrate improved absolute localization and temporal fidelity. These results highlight the value of persistent geometric conditioning for accurate and temporally coherent hand reconstruction from egocentric video.

\textbf{Limitations.} World-space accuracy remains dependent on the estimated camera trajectory and monocular depth prior. Errors in DROID-SLAM rotation, translation, or recovered metric scale propagate to the reconstructed hands and can produce substantial world-space offsets even when camera-space pose is accurate. On ARCTIC, camera MPJPE is 17.54~mm, whereas absolute world MPJPE remains 120.86~mm. \method{} achieves lower absolute world error than the compared baselines, including those using the shared metric-scaled DROID trajectory. Improving camera motion and depth estimation remains necessary for more accurate global hand placement.

\label{end:main}
\label{start:references}
\setlength{\bibsep}{3pt}
\bibliography{references/metrichand_refs}
\bibliographystyle{styles/metrichand}

\clearpage
\end{document}

%% file: tables/crossdit3_main.tex
\begin{table}[t]
\centering\footnotesize
\setlength{\tabcolsep}{1.5pt}
\renewcommand{\arraystretch}{1.10}
\caption{\textbf{Comparison on HOT3D and ARCTIC.} Position errors are in mm and world acceleration error in m/s$^2$ ($\downarrow$). \colorbox{metricFirst}{1st}, \colorbox{metricSecond}{2nd}, and \colorbox{metricThird}{3rd} rank available values within each dataset. Gain compares Ours with the actual second-ranked error.}
\label{tab:main_results}
\label{tab:main_hot3d}
\label{tab:main_arctic}
\begin{tabular*}{\linewidth}{@{\extracolsep{\fill}}cl*{8}{r}@{}}
\toprule
& \multirow{2}{*}{Method} & \multicolumn{4}{c}{Camera} & \multicolumn{4}{c}{World} \\
\cmidrule(lr){3-6}\cmidrule(lr){7-10}
& & {\scriptsize MPJPE$\,\downarrow$} & {\scriptsize\shortstack{RA-\\MPJPE$\,\downarrow$}} & {\scriptsize\shortstack{PA-\\MPJPE$\,\downarrow$}} & {\scriptsize WristErr$\,\downarrow$} & {\scriptsize\shortstack{ABS-W-\\MPJPE$\,\downarrow$}} & {\scriptsize\shortstack{W-\\MPJPE$\,\downarrow$}} & {\scriptsize\shortstack{WA-\\MPJPE$\,\downarrow$}} & {\scriptsize Accel.$\,\downarrow$} \\
\midrule
\multirow{11}{*}{\rotatebox{90}{\textbf{HOT3D}}} & HaMeR-SLAM & 67.87 & 58.23 & 9.57 & 52.29 & 73.28 & 148.23 & 38.43 & 15.91 \\
 & WiLoR-SLAM & 53.78 & 16.58 & 6.70 & 48.89 & 59.21 & 75.90 & 24.65 & 11.39 \\
 & HMP-SLAM & 73.44 & 62.89 & 11.70 & 59.54 & 78.77 & 152.00 & 40.31 & 7.39 \\
 & Dyn-HaMR & 66.06 & 55.46 & 9.19 & 49.98 & 274.08 & 207.97 & 67.98 & 8.85 \\
 & HaWoR & \cellcolor{metricSecond}16.97 & \cellcolor{metricSecond}12.65 & 5.77 & \cellcolor{metricSecond}14.99 & \cellcolor{metricSecond}28.88 & \cellcolor{metricSecond}41.09 & \cellcolor{metricSecond}15.47 & \cellcolor{metricThird}5.57 \\
 & HandFlow & \cellcolor{metricThird}34.89 & \cellcolor{metricThird}14.26 & \cellcolor{metricSecond}5.23 & \cellcolor{metricThird}31.45 & \cellcolor{metricThird}42.59 & 63.27 & 22.56 & 8.63 \\
 & ACE-EgoHand (known-K) & 113.92 & 16.19 & \cellcolor{metricThird}5.31 & 102.06 & 118.43 & \cellcolor{metricThird}55.95 & \cellcolor{metricThird}20.24 & \cellcolor{metricSecond}4.13 \\
 & ACE-EgoHand (K-free) & 118.66 & 15.35 & 5.66 & 110.66 & 122.61 & 98.49 & 31.12 & 6.03 \\
 & MINT & 66.47 & 22.68 & 8.50 & 70.49 & 197.46 & 111.93 & 41.83 & 10.98 \\
 & \textbf{Ours} & \cellcolor{metricFirst}\textbf{8.60} & \cellcolor{metricFirst}\textbf{5.84} & \cellcolor{metricFirst}\textbf{3.18} & \cellcolor{metricFirst}\textbf{7.72} & \cellcolor{metricFirst}\textbf{20.58} & \cellcolor{metricFirst}\textbf{26.71} & \cellcolor{metricFirst}\textbf{9.00} & \cellcolor{metricFirst}\textbf{3.11} \\
 & \textit{Gain vs.\ 2nd} & $49.3\%$ & $53.8\%$ & $39.2\%$ & $48.5\%$ & $28.7\%$ & $35.0\%$ & $41.8\%$ & $24.7\%$ \\
\midrule
\multirow{10}{*}{\rotatebox{90}{\textbf{ARCTIC}}} & HaMeR-SLAM & 54.56 & 25.67 & 9.68 & 56.05 & 166.19 & 83.51 & 31.21 & 17.14 \\
 & WiLoR-SLAM & \cellcolor{metricSecond}36.23 & 14.23 & \cellcolor{metricSecond}6.18 & \cellcolor{metricThird}34.56 & \cellcolor{metricSecond}152.89 & 68.80 & 31.39 & 21.62 \\
 & HMP-SLAM & 95.82 & 32.63 & 11.89 & 100.61 & 199.30 & 99.51 & 37.31 & 9.38 \\
 & Dyn-HaMR & 50.56 & 25.20 & 9.87 & 50.12 & 671.54 & 160.91 & 41.91 & 11.59 \\
 & HaWoR & \cellcolor{metricThird}37.26 & 22.45 & \cellcolor{metricThird}7.10 & \cellcolor{metricSecond}32.23 & \cellcolor{metricThird}154.36 & 68.52 & 28.17 & 12.28 \\
 & ACE-EgoHand (known-K) & 56.57 & \cellcolor{metricSecond}12.04 & 7.30 & 57.56 & 160.71 & \cellcolor{metricSecond}54.66 & \cellcolor{metricThird}24.25 & \cellcolor{metricSecond}6.76 \\
 & ACE-EgoHand (K-free) & 58.70 & \cellcolor{metricThird}12.28 & 7.32 & 60.39 & 163.84 & \cellcolor{metricThird}55.78 & \cellcolor{metricSecond}24.23 & \cellcolor{metricThird}6.84 \\
 & MINT & 116.45 & 27.23 & 9.68 & 106.91 & 621.82 & 107.94 & 45.37 & 13.49 \\
 & \textbf{Ours} & \cellcolor{metricFirst}\textbf{17.54} & \cellcolor{metricFirst}\textbf{11.28} & \cellcolor{metricFirst}\textbf{5.18} & \cellcolor{metricFirst}\textbf{20.02} & \cellcolor{metricFirst}\textbf{120.86} & \cellcolor{metricFirst}\textbf{37.40} & \cellcolor{metricFirst}\textbf{19.08} & \cellcolor{metricFirst}\textbf{4.39} \\
 & \textit{Gain vs.\ 2nd} & $51.6\%$ & $6.3\%$ & $16.2\%$ & $37.9\%$ & $20.9\%$ & $31.6\%$ & $21.3\%$ & $35.1\%$ \\
\bottomrule
\end{tabular*}
\end{table}

%% file: tables/ablation_summary.tex
\begin{table}[t]
\centering\small
\setlength{\tabcolsep}{3pt}
\renewcommand{\arraystretch}{1.10}
\caption{\textbf{Depth representation ablation on ARCTIC.} The first four controls share a metric regression head. Position errors are in mm ($\downarrow$).}
\label{tab:ablation_summary}
\label{tab:ablation_arctic}
\begin{tabular*}{\linewidth}{@{\extracolsep{\fill}}lrrrr@{}}
\toprule
& \multicolumn{3}{c}{Camera} & World \\
\cmidrule(lr){2-4}\cmidrule(lr){5-5}
Configuration & MPJPE & PA-MPJPE & WristErr & ABS-W \\
\midrule
RGB-only regression & {36.34} & {6.78} & {35.27} & {151.74} \\
RGB-D regression (concatenation) & {31.65} & {6.26} & {30.98} & {147.36} \\
RGB-D regression (cross-attention) & {29.64} & {6.01} & {28.56} & {142.14} \\
RGB-D regression (topology-aware) & {19.98} & {5.79} & {23.89} & {127.49} \\
\midrule
Topology-aware conditions + flow & {17.54} & {5.18} & {20.02} & {120.86} \\
\bottomrule
\end{tabular*}
\end{table}

%% file: tables/component_multidataset.tex
\begin{table}[!htb]
\centering\footnotesize
\setlength{\tabcolsep}{1.25pt}
\renewcommand{\arraystretch}{1.10}
\caption{\textbf{Component and geometry-conditioning ablations on HOT3D and ARCTIC.} Position errors are in mm and world acceleration error in m/s$^2$ ($\downarrow$). Controls follow Sec.~\ref{sec:ablations}; Table~\ref{tab:ablation_summary} compares depth representations.}
\label{tab:component_multidataset}
\label{tab:ablation_hot3d}
\begin{tabular*}{\linewidth}{@{\extracolsep{\fill}}l*{5}{r}@{\hspace{7pt}}*{5}{r}@{}}
\toprule
\multirow{3}{*}{Configuration} & \multicolumn{5}{c}{HOT3D} & \multicolumn{5}{c}{ARCTIC} \\
\cmidrule(lr){2-6}\cmidrule(lr){7-11}
& \multicolumn{3}{c}{Camera} & \multicolumn{2}{c}{World} & \multicolumn{3}{c}{Camera} & \multicolumn{2}{c}{World} \\
\cmidrule(lr){2-4}\cmidrule(lr){5-6}\cmidrule(lr){7-9}\cmidrule(lr){10-11}
& {\scriptsize MPJPE} & {\scriptsize\shortstack{PA-\\MPJPE}} & {\scriptsize WristErr} & {\scriptsize\shortstack{ABS-W-\\MPJPE}} & {\scriptsize Accel.}
& {\scriptsize MPJPE} & {\scriptsize\shortstack{PA-\\MPJPE}} & {\scriptsize WristErr} & {\scriptsize\shortstack{ABS-W-\\MPJPE}} & {\scriptsize Accel.} \\
\midrule
\textbf{Full design} & \cellcolor{metricFirst}\textbf{8.60} & \cellcolor{metricFirst}\textbf{3.18} & \cellcolor{metricFirst}\textbf{7.72} & \cellcolor{metricFirst}\textbf{20.58} & \cellcolor{metricFirst}\textbf{3.11} & \cellcolor{metricFirst}\textbf{17.54} & \cellcolor{metricFirst}\textbf{5.18} & \cellcolor{metricFirst}\textbf{20.02} & \cellcolor{metricFirst}\textbf{120.86} & \cellcolor{metricFirst}\textbf{4.39} \\
\midrule
w/o Topology-aware sampling & {17.85} & {4.75} & {12.39} & {31.39} & {4.58} & {38.91} & {6.78} & {35.21} & {154.36} & {5.47} \\
w/o Flow matching & 10.16 & 3.39 & 9.11 & 23.14 & 4.16 & 19.98 & 5.79 & 23.89 & 127.49 & {7.93} \\
w/o Memory queue & 15.42 & 3.98 & 16.55 & 27.78 & 4.05 & 25.67 & 5.96 & 27.37 & 136.24 & 6.31 \\
w/o HOI region context & 19.73 & 4.38 & 18.84 & 31.12 & 4.53 & 32.42 & 5.49 & 28.67 & 139.67 & 5.08 \\
w/o Quality weighting & 11.26 & 3.78 & 10.29 & 21.94 & 3.78 & 25.35 & 5.40 & 23.86 & 132.69 & 4.96 \\
\midrule
\multicolumn{11}{l}{\textit{Geometry conditioning}} \\
Depth for coarse estimation only & {23.76} & {4.97} & {21.33} & {38.49} & {5.93} & {38.95} & {6.56} & {36.04} & {168.93} & {8.19} \\
Geometry at first step only & {15.96} & {4.14} & {17.47} & {29.35} & {4.59} & {28.95} & {5.99} & {28.12} & {141.66} & {6.28} \\
\bottomrule
\end{tabular*}
\end{table}